%% file: main.tex
\documentclass[runningheads]{llncs}

\usepackage{eccv}

\usepackage{eccvabbrv}

\usepackage{graphicx}
\usepackage{booktabs}

\usepackage{multirow}

\usepackage{adjustbox}

\usepackage{bm}

\let\svthefootnote\thefootnote
\newcommand\freefootnote[1]{%
  \let\thefootnote\relax%
  \footnotetext{#1}%
  \let\thefootnote\svthefootnote%
}

\usepackage[accsupp]{axessibility}  

\usepackage{hyperref}

\usepackage{orcidlink}

\begin{document}

\title{Customize-A-Video: One-Shot Motion Customization of Text-to-Video Diffusion Models}

\titlerunning{Customize-A-Video}

\author{
Yixuan Ren$^{1}$\textsuperscript{*}
\and
Yang Zhou$^{2}$
\and
Jimei Yang$^{2}$
\and
Jing Shi$^{2}$
\and
Difan Liu$^{2}$
\and
Feng Liu$^{2}$
\and
Mingi Kwon$^{3}$\textsuperscript{*}
\and
Abhinav Shrivastava$^{1}$
}

\authorrunning{Y.~Ren et al.}

\institute{
$^{1}$ University of Maryland, College Park \quad $^{2}$Adobe Research \quad $^{3}$Yonsei University
}

\maketitle

\freefootnote{\textsuperscript{*} Major work was done during internships at Adobe.}

\input{secs/0_abstract}
\input{secs/1_intro}
\input{secs/2_related_work}
\input{secs/3_method}
\input{secs/4_results}
\input{secs/5_app}
\input{secs/6_conclusion}


%
%
\bibliographystyle{splncs04}
\bibliography{main}

\clearpage

\appendix
\input{secs/X_suppl}

\end{document}

%% file: secs/0_abstract.tex
\begin{abstract}
Image customization has been extensively studied in text-to-image (T2I) diffusion models, leading to impressive outcomes and applications.
With the emergence of text-to-video (T2V) diffusion models, its temporal counterpart, motion customization, has not yet been well investigated.
To address the challenge of one-shot video motion customization, we propose Customize-A-Video that models the motion from a single reference video and adapts it to new subjects and scenes with both spatial and temporal varieties.
It leverages low-rank adaptation (LoRA) on temporal attention layers to tailor the pre-trained T2V diffusion model for specific motion modeling.
To disentangle the spatial and temporal information during training, we introduce a novel concept of appearance absorbers that detach the original appearance from the reference video prior to motion learning.
The proposed modules are trained in a staged pipeline and inferred in a plug-and-play fashion, enabling easy extensions to various downstream tasks such as custom video generation and editing, video appearance customization and multiple motion combination.
Our project page can be found at \url{https://customize-a-video.github.io}.
\keywords{Video Motion Customization \and Text-to-Video Diffusion Models \and Low-Rank Adaptation}
\end{abstract}

%% file: secs/1_intro.tex
\section{Introduction}
\label{sec:intro}

Replicating an iconic motion in novel scenes is highly desirable for video creation.
Recent large-scale diffusion-based text-to-video (T2V) generation models~\cite{wang2023modelscope, chai2023stablevideo} demonstrate impressive outcomes in generating imaginative videos based on text depictions.
However, they struggle with precise motion control and often demand extensive prompt engineering.
Another thread of work on video editing~ \cite{wu2023tune, esser2023structure, wang2023videocomposer, ceylan2023pix2video} leverages large image generative models for appearance alteration, and introduces frame-wise precise controls via DDIM inversion~\cite{song2020denoising, mokady2023null} or ControlNet~\cite{zhang2023adding}. 
While achieving promising motion transfer results with variations in appearance and texture, these methods rigidly adhere to the reference frame structure and layout and fail to provide variability in the motion itself, such as new positions, intensities, camera views, or quantity of subjects.

\begin{figure}[t]
\centering
\captionsetup{type=figure}
\includegraphics[width=\textwidth]{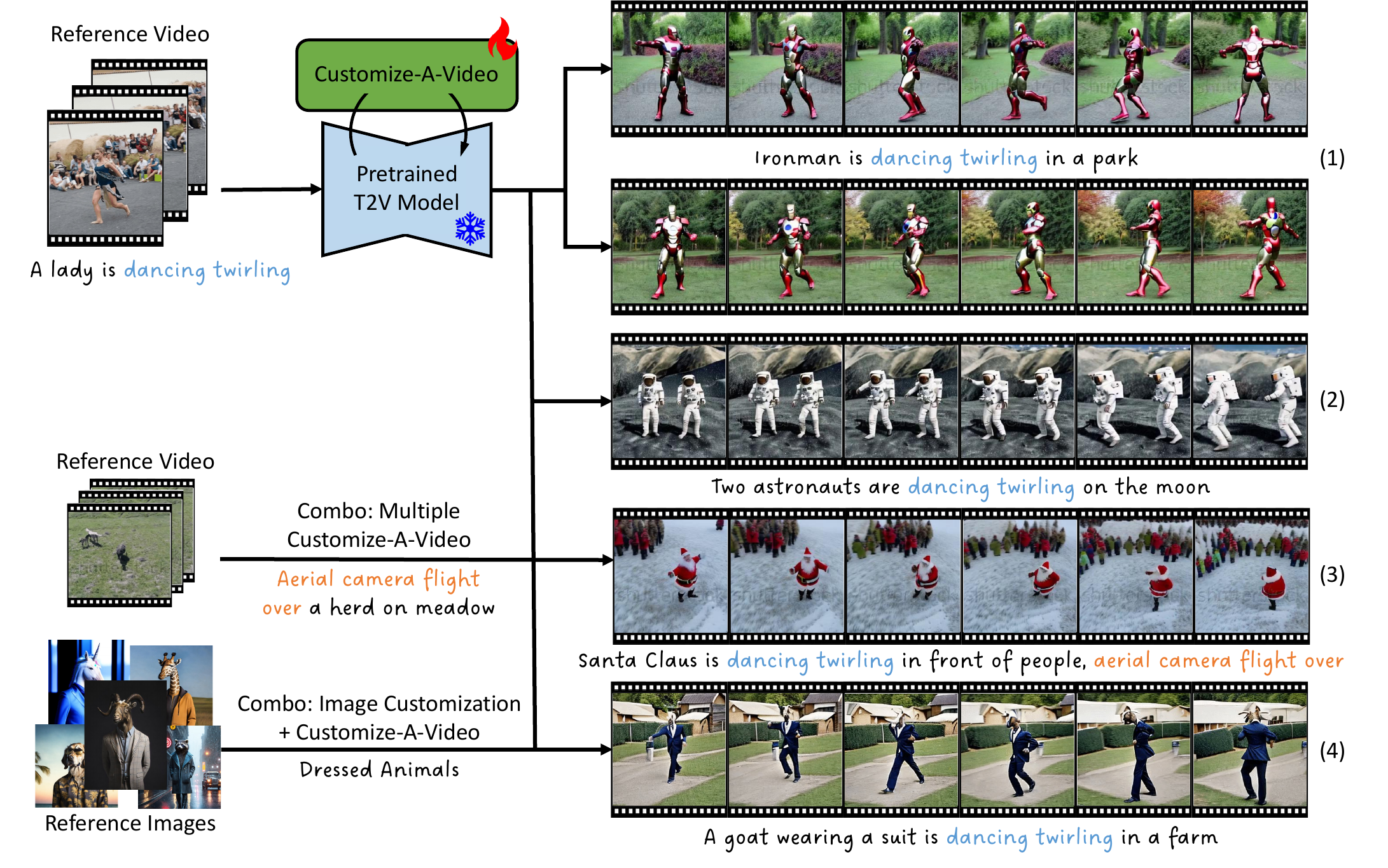}
\captionof{figure}{
    Customize-A-Video takes as input a single reference video (top left) and transfers its motion onto new generated videos with plausible variance.
    \textbf{(1)} Transferring the dancing twirling from the lady onto Ironman with two random output variants.
    \textbf{(2)} Transferring the motion onto multiple subjects.
    \textbf{(3)} Combining multiple motion customization together, i.e., both \textit{dancing twirling} and with \textit{aerial camera flight over}.
    \textbf{(4)} Combining proposed motion customization and existing image customization methods (\cite{Kappa_Neuro} in the example) to support both appearance and motion customization.
}
\label{fig:teaser}
\end{figure}

Image customization of T2I models has been widely explored~\cite{ruiz2023dreambooth,gal2022image} where a specific unique appearance is modeled and composed into novel roles and scenes.
These modules are trained on a small set of images that share the same concept.
They are then able to reproduce the desired concept needless of complex prompt engineering, while also allowing for diversity in poses, views, lighting, etc. compared to direct stitching and editing approaches.
Inspired by this, we introduce a new task of video motion customization and present a novel one-shot method named \textit{Customize-A-Video} (Fig.~\ref{fig:teaser}) built upon T2V diffusion models.
It customizes the pre-trained model with the motion learned from the reference video, enabling it to be easily adapted to new subjects and scenes.
This includes not only precise transfer but also variations in motion intensities, positions, quantity of subjects, and camera views.
These variations make the output videos more dynamic and engaging, as opposed to the robotic rhythm or unnatural appearance of frame-wise tampering.

Specifically, we start from utilizing a common customization technique, Low-Rank Adaptation (LoRA)~\cite{hu2021lora}, applied on a pre-trained T2V diffusion model~\cite{wang2023modelscope} to capture the motion signature in the reference video.
Applying LoRA directly to the entire T2V models proves less effective in motion preservation, as spatial and temporal characteristics are intricately entangled and both will be learned simultaneously.
Therefore, we apply LoRA only on temporal cross-frame attention layers, creating Temporal LoRA (T-LoRA), which is more concentrated on capturing motion dynamics from the video.
In comparison to other popular customization algorithms, LoRA also offers a portable model size and requires minimal training data, as well as the simplicity of plug-and-play for easy extensibility to collaborate with additional customization modules.

While LoRA works well on few-shot customization tasks through the residual module weights, a portion of spatial features still leak into it when trained on a single reference video.
Concurrent efforts attempt to address this challenging yet significant issue by either demanding a small dataset with diverse appearances and the consistent motion ~\cite{materzynska2023customizing, wei2023dreamvideo}, or stopping training early and supplementing the underfit temporal modules with direct control signals from the reference video~\cite{jeong2023vmc}.
To tackle this issue and facilitate one-shot video customization, we introduce an innovative Appearance Absorber module to further decompose static signals from dynamics.
The key idea of this module is to \textit{absorb} the appearance out of the reference video, leaving only the desired motion information for the Temporal LoRA to model.

We introduce a staged training and inference pipeline as illustrated in Fig.~\ref{fig:archs} to connect all the components we have proposed while keeping them independent.
In the first stage, we build and train the appearance absorber on unordered reference video frames to capture frame-wise spatial information, such as the subject appearance and the background scene.
In the second stage, we load the trained appearance absorber in frozen state, and construct the Temporal LoRA on the temporal layers of the T2V model to train.
The appearance absorber has encoded the static frames and therefore helps the Temporal LoRA focus primarily on temporal signals, minimizing the spatial information leakage into motion customization modules.
During the inference stage, we remove the appearance absorber and load solely the trained Temporal LoRA. Given a text prompt containing novel subjects and scenes, our model not only accurately transfers the learned motion signature to the new appearance, but also produces diverse motions in terms of their intensities, positions, and camera views.

To summarize, our contributions involve:
\begin{itemize}
\item We present a novel one-shot motion customization method for single reference video based on pre-trained text-to-video diffusion models;
\item We introduce Temporal LoRA to learn the motion from a single reference video, facilitating motion transfer with not only accuracy but also variety;
\item We propose the general class of Appearance Absorbers to dedicatedly decompose the spatial information out of the reference video, effectively excluding it from the motion customization process;
\item Our modules feature the plug-and-play and staged fashion and can be smoothly extended to various downstream applications.
\end{itemize}

%% file: secs/2_related_work.tex
\section{Related Work}
\label{sec:related}

\paragraph{Text-to-Video Generation Models.}
Text-to-video (T2V) generation task generates videos from given text prompts specifying the expected appearances and motions.
It has been widely explored previously using GANs~\cite{pan2017create, li2018video, kim2020tivgan} and transformers~\cite{villegas2022phenaki, wu2022nuwa, yan2021videogpt, hong2022cogvideo, ge2022long}.
With the boost of T2I diffusion models, T2V diffusion models become subsequently under fast development.
\cite{khachatryan2023text2video} reprogram the 2D spatial attentions into 3D temporal attentions to handle the new temporal dimension.
\cite{ho2022video, ho2022imagen, singer2022make, zhou2022magicvideo, blattmann2023align, guo2024animatediff, wang2023modelscope, liu2023dual, chen2023videocrafter1} insert spatio-temporal 3D convolutions and/or cross-frame attentions to regulate the output temporal consistency from the random input noise.
\cite{luo2023videofusion, ge2023preserve, khachatryan2023text2video, liu2023dual} design explicitly disentangled noise prior between key frames and residues to enforce temporal coherency.
T2V models designate the generated content through text prompts, demanding significant engineering effort to prompt it to produce desired motions  in details.

\paragraph{T2I-based Video Editing.}
Leveraging the control signal directly from a reference video by editing it into new appearances is an efficient practice to precisely transfer the motion and has been studied by various methods.
\cite{wu2023tune, wang2023zero, zhang2023towards, qi2023fatezero, geyer2023tokenflow, shin2023edit, liu2023video} leverage the inverse denoising process or degradation of the reference video frames to maintain the desired motion while altering its appearance through T2I generation.
\cite{chu2023video, esser2023structure, wang2023videocomposer, zhang2023controlvideo, liew2023magicedit, ma2023follow, AILab-CVC} adopt controllable image generation approaches~\cite{zhang2023adding, mou2023t2i} and extract the low-level reference signals such as their depth or edge maps to guide the generation process.
\cite{zhao2023make, ceylan2023pix2video, chen2023control, zhao2023controlvideo, yang2023rerender} make use of the combination of above techniques.
However, such methods fall short as they focus more on adopting novel appearances, and merely duplicate the original motion exactly but with no temporal diversity to vary in the motion intensity and velocity, subject position and quantity etc.

\paragraph{Video Motion Customization.}
Model customization is the task of adapting the original output to a new specific domain by adjusting the pre-trained model weights.      
It was first introduced for T2I models to personalize in spatial aspects such as identity, art style or pose~\cite{ruiz2023dreambooth, gal2022image,kumari2023multi}.      
Recently, the idea of customizing the motion given reference videos has also been emerging and evolving rapidly.
\cite{wu2023tune, shin2023edit, liu2023video, zhang2023towards} add temporal attentions from scratch on pre-trained T2I models and finetune them on a single video.
Concurrent work~\cite{materzynska2023customizing, jeong2023vmc} tunes the temporal layers in place in a pre-trained T2V model with either a regularization set or a frame residual loss to reduce the impact of training videos' appearances.
Instead, our method appends residual weights to the original model using LoRA~\cite{hu2021lora} and enables ligthweight training strategy and flexible inference utility.

\cite{molad2023dreamix} represents the first attempt to finetune the spatial and temporal attentions independently for the appearance and motion of a reference video, which have however entangled inference.
Concurrent work \cite{zhang2023motioncrafter} employs two parallel UNets and tune one of them with an appearance normalization loss to disentangle the motion as well as maintain the generalization ability on new appearances.
Another concurrent work~\cite{wei2023dreamvideo} adds adapters conditioned on one frame to decompose pure motion from its appearance, requiring additional input of an image when inference while ours asks for the minimal input of text prompt only.
Concurrent work~\cite{zhao2023motiondirector} applies dual-path LoRAs and trains them jointly with an appearance-debiased loss.
In contrast, our approach adopts a staged training pipeline with independent tuning configurations, where our appearance absorber class can be easily extended to more candidates than LoRA, or reusing third-party modules pre-trained on in-the-wild images or videos.

%% file: secs/3_method.tex
\begin{figure}[t]
  \centering
  \includegraphics[width=\linewidth]{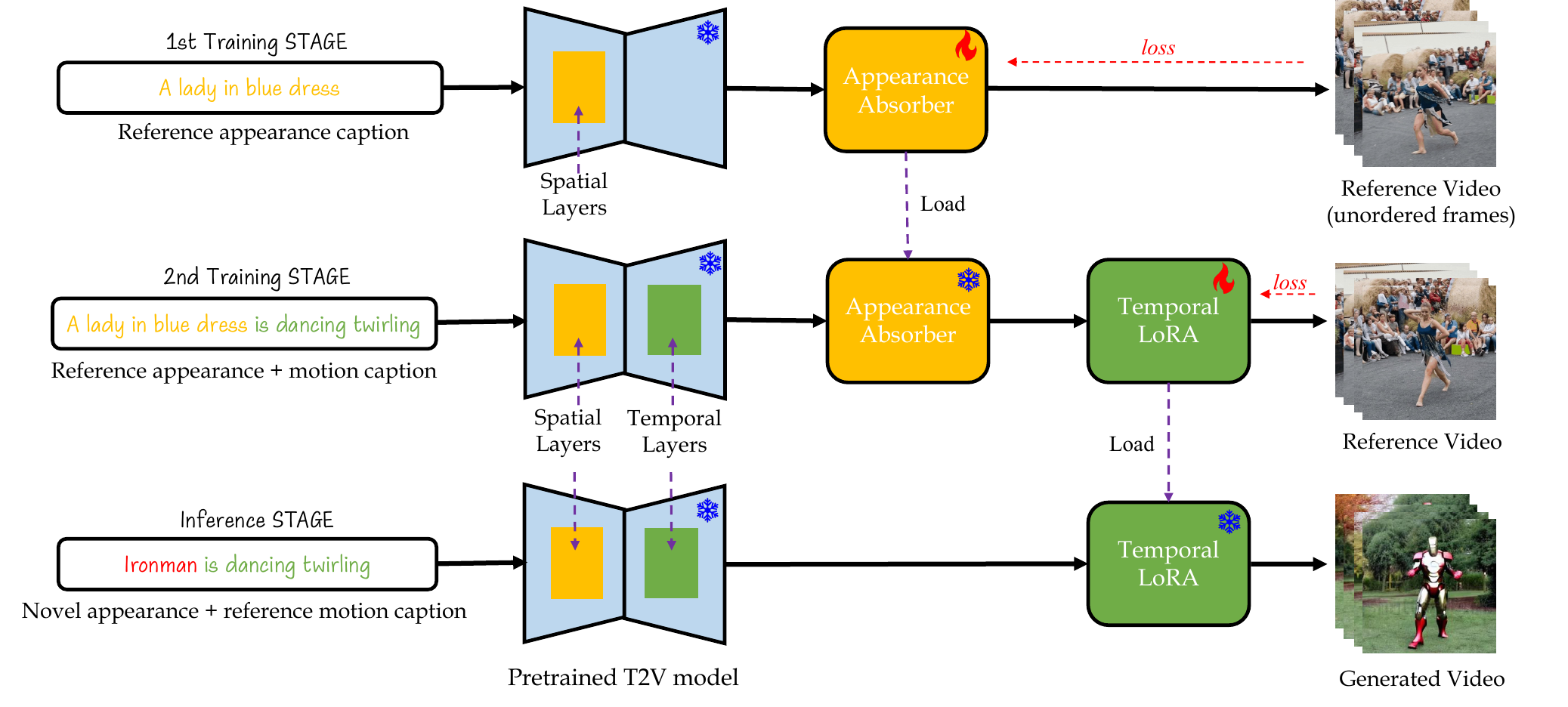}
  \caption{
    Our Temporal LoRA, Appearance Absorbers and their training and inference processes.
    All noise and denoising schedules are omitted for simplicity.
    \textbf{(1)} We bypass all temporal layers in a base T2V diffusion model and apply appearance absorber such as S-LoRA or Textual Inversion on its spatial attention layers. The module is trained on unordered video frames.
    \textbf{(2)} We apply T-LoRA on all temporal attentions in the full base T2V model. The trained appearance absorber is also loaded and frozen. The module is trained on the target video.
    \textbf{(3)} During inference, only the trained T-LoRA is loaded. A new video with the customized motion is generated by a prompt describing the new appearance and the target motion.
    }
  \label{fig:archs}
\end{figure}

\section{Method}
We present a novel motion customization method based on pre-trained T2V diffusion models for a single reference video. We suggest learning the motion concept from the reference video through a LoRA module designed for temporal layers of the T2V model. Given the challenging nature of working with a single datum, we develop a staged training strategy with an appearance absorber module to disentangle spatial information from motion.
Fig.~\ref{fig:archs} shows an illustration of each proposed module and its connection to the base T2V model.

\subsection{Preliminary}

\paragraph{Text-to-Video Diffusion Models.}
A text-to-video (T2V) diffusion model trains a 3D UNet to generate videos in a series denoising steps conditioned on a input text prompt.
The 3D UNet usually consists of spatial self- and cross-attentions, 2D and 3D convolutions, and temporal cross-frame attentions.
Given the $F$ frames $x^{1...F}$ of a video, the 3D UNet $\theta$ is trained by minimizing
\begin{equation}
    L_{\theta} = \mathbb{E}_{x^{1...F}, \epsilon, t } [ \| \epsilon - \epsilon_{\theta}(x_{t}^{1...F}, t, \tau_{v}(y)) \| ]
\label{eq:diffusion}
\end{equation}
at every denoising step $t=T,...,0$. $\epsilon$ is Gaussian noise and $\epsilon_{\theta}$ is the UNet prediction, $\tau$ is the text encoder with token sets $v$, and $y$ is the text prompt.

\paragraph{Low-Rank Adaptation.}
Low-Rank Adaptation (LoRA)~\cite{hu2021lora} was proposed for adapting pre-trained large language models to downstream tasks.
It has also been widely developed for image customization models.
LoRA applies a residue path of two low-rank matrices $\theta_{B}\in\mathbb{R}^{d\times r}, \theta_{A}\in\mathbb{R}^{r\times k}$ on an attention layer, whose original weight is $\theta_{0}\in\mathbb{R}^{d\times k}$, $r\ll\min(d, k)$.
The new forward path is
\begin{equation}
    \theta = \theta_{0} + \alpha \Delta \theta = \theta_{0} + \alpha \theta_{B} \theta_{A}.
\end{equation}
where $\alpha$ is a coefficient adjusting the strength of the added LoRA.

\subsection{Customize-A-Video}

We proposed two critical modules to customize the pre-trained T2V model for a single reference video. \textit{Temporal LoRA} is introduced to learn motion from a reference video, whereas \textit{Appearance Absorbers} are crafted to improve the separation of spatial and temporal information within the single reference video.

\subsubsection{Temporal LoRA}

Inspired by \cite{hu2021lora}, we introduce Temporal LoRA (T-LoRA), a technique for capturing motion characteristics from input videos and enabling motion customization for new appearance via text prompts.
We apply LoRAs on all temporal cross-frame attention layers of the base T2V model~\cite{wang2023modelscope} to maximize modeling motion signals.
Our ablation studies reveal that T-LoRA outperforms applying LoRA to other non-temporal attention layers, as T-LoRA targets at preserving motion while discarding unnecessary input appearance (see Sec.~\ref{sec:ablations}).

\subsubsection{Appearance Absorbers}

To separate spatial signals from temporal signals within a single video, we abstract the general class of Appearance Absorbers.
Its objective is to absorb the spatial information, including the identity, texture, scene, etc., out of the training video, such that the reference motion can be exclusively modeled by our T-LoRA.
To achieve this, we construct the absorbers leveraging a set of image customization modules, and use them in an \textit{inverse} manner compared to their original role:
\begin{equation}
\begin{aligned}
    \textrm{Image Model} \, \bm{+} \, \textit{\textrm{Image Customization Module}} & \rightarrow \textrm{Image Custom Model} ; \\
    \underbrace{\textrm{Video Model} \, \bm{-} \, \textit{\textrm{Appearance Absorber}}}_\text{\textit{remove spatial signals from training video}} \, + \, \textrm{T\text{-}LoRA} & \rightarrow \textrm{Video Custom Model} .
    \notag
\end{aligned}
\end{equation}
Our appearance absorbers can be built upon including but not limited to:

\begin{itemize}

\item \textit{Spatial LoRA.}
We apply LoRA on only the spatial attention layers in a T2V model to adopt solely the spatial information out of the video frames.
LoRA modules are injected in all self-attention layers of the frames and cross-attention layers between frames and the text prompt.
We call it spatial LoRA (S-LoRA) to distinguish from our T-LoRA for temporal modeling.

\item \textit{Textual Inversion.}
We utilize textual inversion~\cite{gal2022image} as another approach to collect spatial features from the training video.
It creates learnable placeholder tokens, initialized with briefly depicting words of the video appearance, to assimilate relevant spatial information through the text tokenizer.

\end{itemize}

These image customization modules are adept at modeling appearance signals from limited number of frames of a single video in a few-shot manner, and thus we prefer less finetuning-based customization methods such as \cite{ruiz2023dreambooth} since they require a considerable amount of training and regularization data.
All types of appearance absorbers can be employed individually or jointly.

\subsubsection{Training and Inference Pipelines}

Our motion customization pipeline in Fig.~\ref{fig:archs} consists of two training stages for appearance absorbers and T-LoRA respectively, and one inference stage to finally generate output videos with novel text prompts.
Our configurable pipeline has dedicated stage for each module and is universal for extensive types of its components.

\paragraph{First training stage.} 
We train appearance absorber modules first.
Since they are originated from T2I models, we propose to specially train them by bypassing all temporal layers in the T2V model, including temporal attention layers and 3D convolution layers in the denoising UNet.
We train them with the appearance description $y_{S}$ cut out of the full caption so that they focus on learning the spatial information.
The training images are unordered frames of the reference video.
We follow their native loss as in \cite{hu2021lora, gal2022image} to train each type of appearance absorber.
Formally, for S-LoRA $\Delta\theta_{S}$:
\begin{equation}
    L_{\Delta\theta_{S}} = \mathbb{E}_{x, \epsilon, t } [ \| \epsilon - \epsilon_{\theta_{0}+\Delta\theta_{S}}(x_{t}^{f}, t, \tau_{v_{0}}(y_{S})) \| ],
\end{equation}
and for textual inversion $\Delta v$:
\begin{equation}
    L_{\Delta v} = \mathbb{E}_{x, \epsilon, t } [ \| \epsilon - \epsilon_{\theta_{0}}(x_{t}^{f}, t, \tau_{v_{0}+\Delta v}(y_{S})) \| ].
\end{equation}

\paragraph{Second training stage.}
We inject above trained appearance absorbers into the T2V model and maintain their frozen state.
Our T-LoRA is meanwhile injected into the temporal attention layers of the T2V model.
It is trained with the reference video and full ground truth caption consisting of both motion verbs and appearance nouns, by which the appearance absorber is also triggered to yield spatially customized content in static frames.
We train T-LoRA $\Delta\theta_{T}$ using the standard reconstruction loss as in diffusion models~\cite{rombach2022high}:
\begin{equation}
    L_{\Delta\theta_{T}} = \mathbb{E}_{x^{1...F}, \epsilon, t } [ \| \epsilon - (\epsilon_{\theta'+\Delta\theta_{T}}(x_{t}^{1...F}, t, \tau_{v'}(y))) \| ]
\end{equation}
where $\theta'=\theta_{0}+\Delta\theta_{S}$ and $v'=v_{0}+\Delta v$ if respective AAs are employed.

\paragraph{Inference stage.}
During the final inference stage, solely the trained T-LoRA is loaded onto the base T2V model.
Given a new text prompt depicting the reference motion with new appearances and scenes, the customized model generates novel videos animated by the desired motion following the standard denoising process.           
As a result of the customized weights in T-LoRA, our output video transfers the reference motion faithfully as well as with diversity in motion intensities, positions, and camera views etc.

%% file: secs/4_results.tex
\begin{figure}
  \centering
  \includegraphics[width=\linewidth]{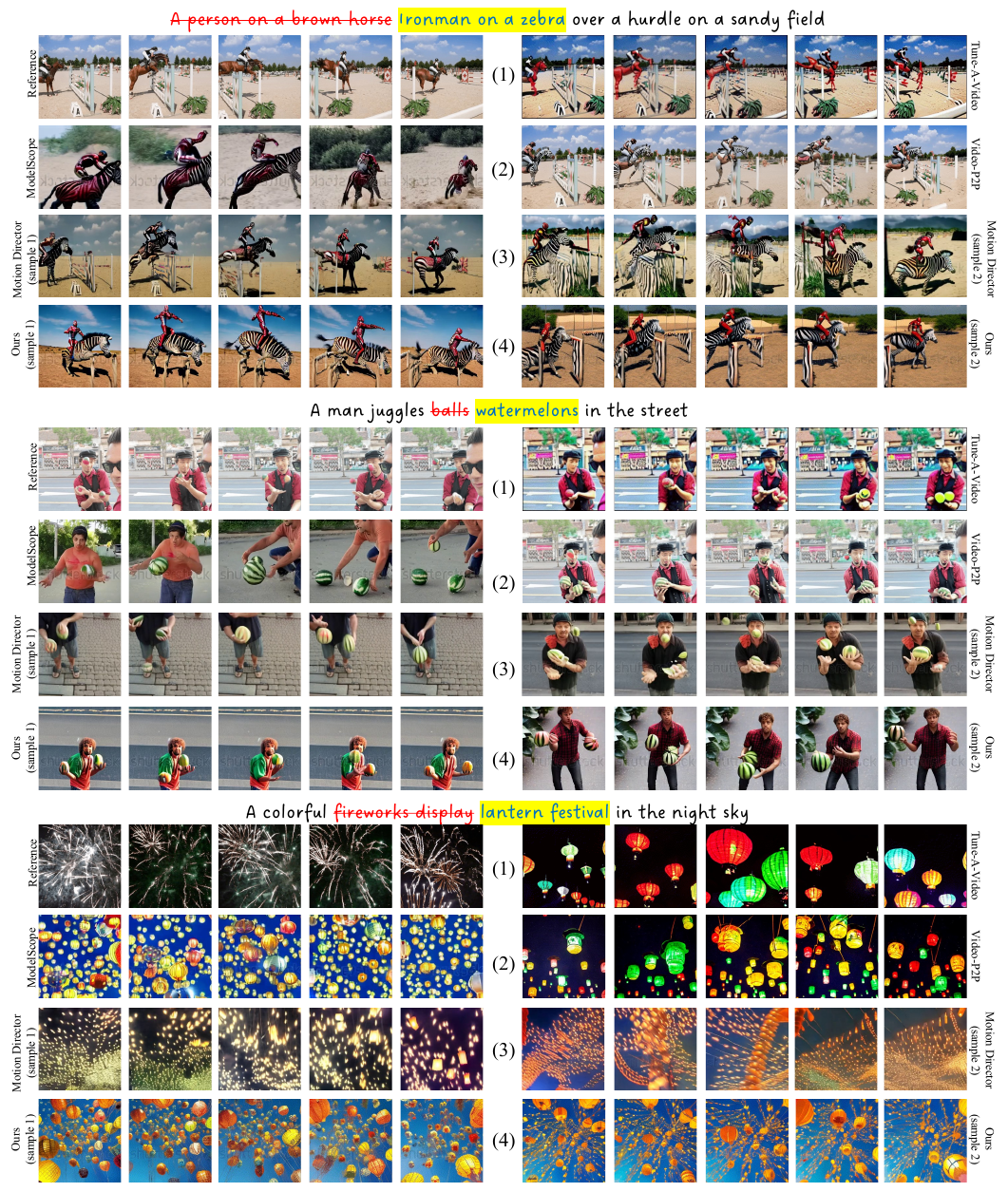}
  \caption{
  Results of one-shot motion customization.
  \textbf{(1-left)} Reference video.
  \textbf{(2-left)} ModelScope~\cite{wang2023modelscope} fails to transfer the reference motion faithfully with only text guidance.
  \textbf{(1-right \& 2-right)} Tune-A-Video~\cite{wu2023tune} and Video-P2P~\cite{liu2023video} rely on DDIM inverted latent input and duplicate the original frame structure deterministically.
  \textbf{(3)} Concurrent work MotionDirector~\cite{zhao2023motiondirector} also generates various output following the reference motion while there exist some appearance and motion artifacts especially for hard examples with complex or intensive movements.
  \textbf{(4)} Our methods generate motion with both accuracy and variety in details such as view perspective and frame layout.
  Two variants generated with random noise are shown for MotionDirector and Ours.
  }
  \label{fig:main}
\end{figure}

\section{Experiments}

\paragraph{Base T2V models.}
Our methods are applicable to general T2V diffusion models. In the following experiments, we hire the ModelScope T2V model~\cite{wang2023modelscope} as the pre-trained base model.
All videos are pre-processed and generated for 2 seconds, 8 FPS and $256\times 256$ resolution.
Training hyperparameters, model size statistics and time consumption analyses are detailed in the supplementary material.

\paragraph{Datasets.}
We select videos from mixed sources, including LOVEU-TGVE-2023~\cite{wu2023cvpr}, WebVid-10M~\cite{Bain21} and DAVIS~\cite{Perazzi2016} datasets to evaluate our method.
\cite{wu2023cvpr} provides ground truth captions and target editing prompts, while we create those for videos from other sources.
We also apply our method on in-the-wild videos and demonstrate its generalization ability.

\paragraph{Comparison methods.}
As of a new task of one-shot video motion customization, we mainly compare to Tune-A-Video~\cite{wu2023tune} and Video-P2P~\cite{liu2023video}, 
which append raw temporal layers to pre-trained T2I models and finetune them on a single reference video.
It is worth noting that they additionally rely on DDIM inverted reference video latent as the input during inference and thus only produce temporally deterministic videos with fixed frame layout and view angle, so we also evaluate their variants removing this condition.
We also compare our method against the pre-trained T2V model, i.e. ModelScope, to prove that our method enhances the base foundation model to produce faithful motions following the reference video that are not trivial to depict via prompt engineering.
Besides, we run the concurrent work MotionDirector~\cite{zhao2023motiondirector} using their released training code with the same configuration as ours, including the same base T2V model and LoRA hyperparameters for fair comparison.

\paragraph{Quantitative metrics.}
We measure the performance quantitatively over a subset containing 53 videos out of \cite{wu2023cvpr} of 2-3 seconds with standard original and editing captions.
We consider comparisons in terms of three metrics:
\textbf{text alignment} between the generated video frames and the inference prompt gauges both generated appearance and motion accuracy, in the form of CLIPScore~\cite{hessel2021clipscore} that associates text and image in a unified space;
\textbf{temporal consistency} between consecutive frames of the generated video indicates the generated motion quality, in the form of LPIPS~\cite{zhang2018unreasonable} that measures deep feature distance;
\textbf{diversity} among multiple generated videos with the same prompt and different random noise involves both spatial and temporal diversity by collating aligned frames at the same timestamp, in the form of LPIPS.
It is calculated on 4 random samples per reference video.

\paragraph{User Study.}
We conduct a human user study among five algorithms with stochastic output: ModelScope, Tune-A-Video and Video-P2P without DDIM inverted latent, our method with both S-LoRA and textual inversion as the appearance absorbers, and MotionDirector.
Every participant is presented 10 random reference videos from \cite{wu2023cvpr} and their output videos.
Each algorithm outputs two videos, and participants are asked to assess their motion fidelity and motion diversity respectively, from 1 (worst) to 5 (best) stars.
Details of the questionnaire design is provided in the supplementary materials.

\begin{table}[t]
  \caption{
  Quantitative comparisons on \cite{wu2023cvpr} dataset.
  $\mathit{\sim}$ \textit{w/o DDIM Inversion} represents the above method without DDIM inverted latent input.
  Video-P2P outputs video clips of 4 FPS with $512\times 512$ resolution.
  MotionDirector is a concurrent work to ours and is tested with either the same LoRA rank or comparable amount of parameters to ours.
  }
  \centering
  \begin{tabular}{@{}lccc@{}}
    \toprule
    Method & Text Alignment$\uparrow$ & Temporal Consistency$\downarrow$ & Diversity$\uparrow$ \\
    \midrule
    ModelScope~\cite{wang2023modelscope} & $31.705$ & $0.175$ & $\boldsymbol{0.636}$ \\
    \midrule
    Tune-A-Video~\cite{wu2023tune} & $31.149$ & $0.185$ & - \\
    $\sim$ w/o DDIM Inversion & $30.304$ & $0.206$ & $0.348$ \\
    Video-P2P~\cite{liu2023video} & $31.001$ & $\underline{0.162}$ & - \\
    $\sim$ w/o DDIM Inversion & $30.876$ & $0.251$ & $0.469$ \\
    \midrule
    MotionDirector~\cite{zhao2023motiondirector} & $\underline{32.500}$ & $0.163$ & $0.606$ \\
    $\sim$ w/ comparable \#params & $31.842$ & $0.166$ & $0.595$ \\
    \midrule
    Ours No AA & $31.687$ & $0.166$ & $0.613$ \\
    Ours S-LoRA AA & $31.913$ & $0.163$ & $0.618$ \\
    Ours TextInv AA & $\boldsymbol{32.632}$ & $\boldsymbol{0.160}$ & $0.621$ \\
    Ours Dual AA & $32.193$ & $0.164$ & $\underline{0.631}$ \\
    \bottomrule
  \end{tabular}
  \label{tab:numbers}
\end{table}

\subsection{Motion Customization from Single Video}

\paragraph{Qualitative Results.}
Fig.~\ref{fig:main} illustrates the comparative visual results of one-shot motion customization.
The base ModelScope T2V model fails to accurately replicate the specific motions as reference guided by simply the text prompt.
On the other hand, Tune-A-Video~\cite{wu2023tune} and Video-P2P~\cite{liu2023video} leverage DDIM inverted latents extracted from reference videos and produce temporally deterministic output with structural constraints by the reference frame layouts.
In contrast to both of them, our approach demonstrates the capability to transfer the reference motion to new scenarios and subjects while introducing temporal variations via random noise input.
Our outcomes not only exhibit diverse subject appearances and background scenes but also showcase variability in motion attributes such as action range, intensity, velocity, and camera perspective.
Concurrent MotionDirector~\cite{zhao2023motiondirector} is also able to generate adapted output with variety, while for the hard examples with complex or intensive motion, such as the juggling and firework in Fig.~\ref{fig:main}, it yields less competitive visual quality than ours.
Our T-LoRA is always trained with well-optimized AAs, while MotionDirector might have appearances leaked into the temporal module when their spatial path is jointly being tuned.
We also notice that real-world videos usually have divergent spatial and temporal complexity, and our dedicated tuning procedures with independent steps and other schedules for each module reach their individual optimal.

\paragraph{Quantitative Results.}
The quantitative results and comparisons are listed in Tab.~\ref{tab:numbers}.
Our methods outperform the base ModelScope~\cite{wang2023modelscope}, Tune-A-Video~\cite{wu2023tune}, Video-P2P~\cite{liu2023video} and concurrent MotionDirector~\cite{zhao2023motiondirector} on both text alignment and temporal consistency.
ModelScope~\cite{wang2023modelscope} provides the highest comprehensive diversity as a foundation model and loses in motion fidelity with text guidance only.
Our methods sacrifice it subtly to gain significant improvements in faithfully customizing the exemplar motion, as well as retaining rich varieties in motion details.
Tune-A-Video~\cite{wu2023tune} and Video-P2P~\cite{liu2023video} have the minimal diversity as strictly constrained by frame structures, although at the cost of which they can achieve acceptable temporal consistency.
MotionDirector~\cite{zhao2023motiondirector} shows comparable text alignment and temporal consistency to ours but its diversity falls behind.

\paragraph{User Study.}
We involved 20 evaluators participating our user study and collected 153 and 156 valid ratings per algorithm on each benchmark.
The averaged scores are listed in Tab.~\ref{tab:userstudy}.
Our method leads on both motion fidelity and motion diversity.
We asked users to rate pure motion fidelity and diversity which are hard to assess by automatic metrics in need of generic motion representations irrespective to spatial structures.
These results are complementary to the text alignment and diversity measured in Tab.~\ref{tab:numbers} that mix spatial and temporal signals.

\begin{table}
  \caption{
  Human user study results on \cite{wu2023cvpr} dataset.
  Methods are evaluated from 1 (worst) to 5 (best) stars on each benchmark.
  }
  \centering
  \begin{tabular}{@{}lcc@{}}
    \toprule
    Method & Motion Fidelity$\uparrow$ \quad & Motion Diversity$\uparrow$ \quad \\
    \midrule
    ModelScope~\cite{wang2023modelscope} & $2.03$ & $2.97$ \\
    Tune-A-Video~\cite{wu2023tune} w/o DDIM Inversion & $2.29$ & $2.23$ \\
    Video-P2P~\cite{liu2023video} w/o DDIM Inversion & $2.29$ & $2.01$ \\
    \midrule
    MotionDirector~\cite{zhao2023motiondirector} & $\underline{3.33}$ & $\underline{3.50}$ \\
    \midrule
    Ours (w/ dual AA) & $\boldsymbol{3.72}$ & $\boldsymbol{3.72}$ \\
    \bottomrule
  \end{tabular}
  \label{tab:userstudy}
\end{table}

\begin{figure}[t]
  \centering
  \includegraphics[width=\linewidth]{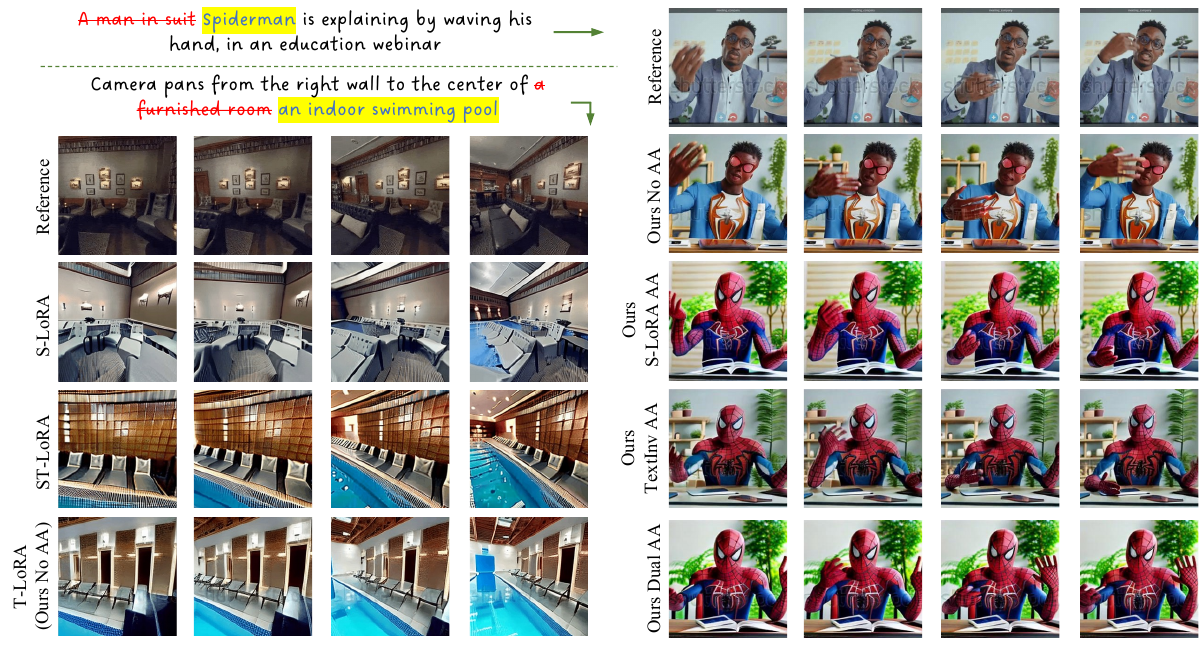}
  \caption{
    \textbf{Left}:
    Ablations on applying LoRAs on different attention layers.
    S-LoRA memorizes the indoor furniture and wall decorations and T-LoRA converts paintings to entrances and sofas to pool benches.
    \textbf{Right}: Ablations on training T-LoRAs with different types of appearance absorbers.
    No AA adds stylish glasses and the logo but remains most the original appearance.
    S-LoRA AA and TextInv AA significantly boost the quality while resulting in the strips on the wall and the partially white sleeves.
    Dual AA reaches best spatial clearance with clear costume and background.
  }
  \label{fig:ablations}
\end{figure}

\subsection{Ablations Studies}
\label{sec:ablations}

\paragraph{LoRAs on Non-temporal Attentions.}
While it is intuitive to apply LoRA on only temporal attention layers to learn video motions without original appearance, we also validate the effects of applying LoRA on the spatial attentions only (\textit{S-LoRA}), or on both spatial and temporal attentions (\textit{ST-LoRA}) in the base T2V model.
Fig.~\ref{fig:ablations} left displays the visualizations in which adding LoRA to spatial attentions significantly impairs the motion modeling.    
The models with spatial elements primarily memorize the video by its spatial layout, resulting in a substantial degradation of both appearance and motion adaptation.

\paragraph{Comparisons among Appearance Absorber Types.}
We explore four different configurations of Appearance Absorbers (AA). \textbf{No AA}: no appearance absorber is used, \textbf{S-LoRA AA}: a spatial LoRA based appearance absorber is used, \textbf{TextInv AA}: a textual inversion based appearance absorber is used, and \textbf{Dual AA}: two appearance absorbers of both above types are used.
The comparison results are unveiled by Fig.~\ref{fig:ablations}.
No AA remains some original appearance in addition to modeling the motion.
S-LoRA AA and TextInv AA are both able to capture the pure action with minimal appearance leakage.
We notice that S-LoRA AA is easier to overfit and sometimes causes spatial artifacts while TextInv AA might tend to underfit and leave spatial residues on the other hand.
We attribute these properties to the spatial structure of S-LoRA weights inside U-Net blocks while textual inversion works via a 1D learnable embedding as new tokens.
Dual AA unites their advantages and leads to a comprehensive result with both the reference motion and new appearance clearly reflected.

%% file: secs/5_app.tex
\begin{figure}[t]
  \centering
  \includegraphics[width=\linewidth]{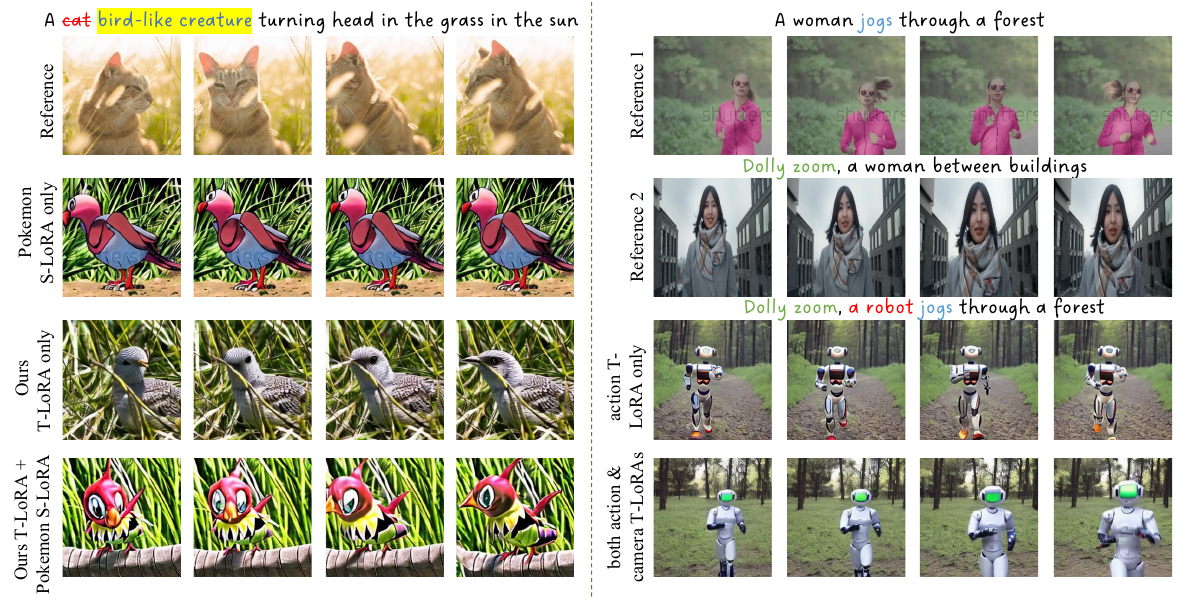}
  \caption{
    \textbf{Left}:
    Video appearance customization with both T-LoRA and existing pre-trained S-LoRA (from~\cite{tungdop2}).
    \textbf{Right}:
    Multiple motion combination with two T-LoRAs loaded at the same time.
    When the robot is slowly jogging, it fast zooms in while the background trees rapidly zoom out (dolly zoom).
  }
  \label{fig:applications}
\end{figure}

\section{Applications}

With the plug-and-play nature of LoRA and our staged training pipeline, we present four downstream applications that demonstrate the collaborative potential of our proposed modules.

\subsubsection{Video Appearance Customization}
Our motion customization module works on temporal layers and thus can cooperate with image customization approaches to manipulate both the temporal and spatial layers in the base T2V model at the same time.
In Fig.~\ref{fig:applications} left, we inject a T-LoRA to present the reference action as well as an image spatial LoRA to reflect the comic style in one comprehensive output.

\subsubsection{Multiple Motion Combination}
T-LoRA is applied to the original layers with residual connections.
Therefore we can customize the base model with multiple T-LoRA modules trained on different source videos to integrate assorted motions into one outcome.
Fig.~\ref{fig:applications} right demonstrates that our method merges the human action of jogging and camera movement of dolly zoom into one target scenario using two T-LoRA modules.

\subsubsection{DDIM Inverted Latent Input}
Our method can easily incorporate additional deterministic controls to perform precise video editing.
The comparison results to Tuna-A-Video\cite{wu2023tune} are shown as Fig.~\ref{fig:ddiminv_external} left.
Our models prove to be also able to benefit from the DDIM inverted latent of the reference video and yield output that reproduces the exact original frame structures.

\subsubsection{Third-Party Appearance Absorbers}
\begin{figure}[t]
  \centering
  \includegraphics[width=\linewidth]{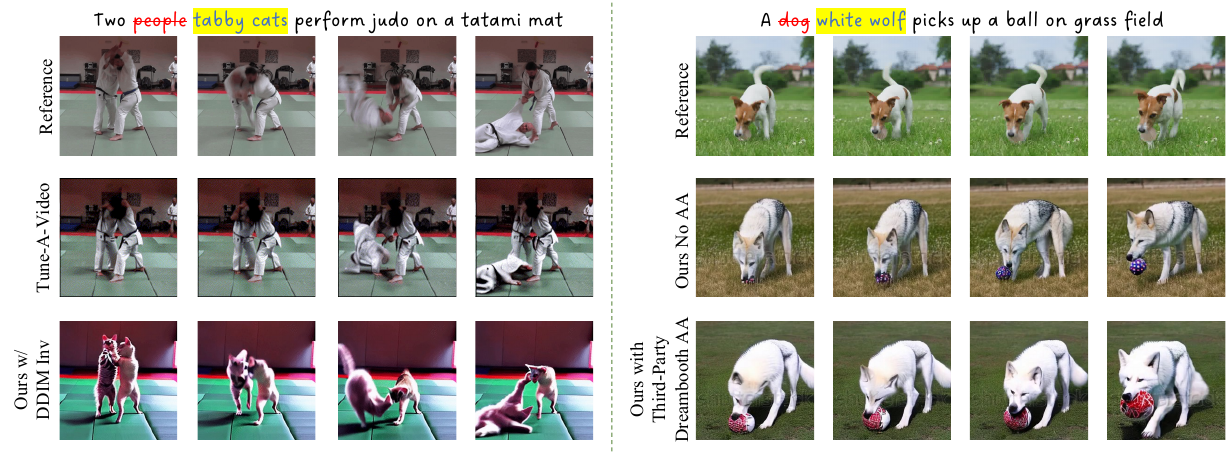}
  \caption{
    \textbf{Left}:
    Precise frame-wise video editing with DDIM inverted latent as the inference input noise.
    \textbf{Right}:
    A third-party Dreambooth UNet pre-tuned on in-the-wild images of the same subject serves as the appearance absorber to train the T-LoRA.
  }
  \label{fig:ddiminv_external}
\end{figure}

Our staged training pipeline enables reusing appearance absorbers across videos or loading third-party image customization modules as ready appearance absorbers when they share the similar appearance.
This skips the first training stage and extends appearance absorber categories to those demand more training data.
In Fig.~\ref{fig:ddiminv_external} right, we finetune the spatial layers of the UNet following Dreambooth~\cite{ruiz2023dreambooth} and Eq.~\ref{eq:diffusion} on other photos of the same dog.
Our T-LoRA trained with it avoids the leakage of the dog color to the white wolf, compared to that trained with the original base model.

%% file: secs/6_conclusion.tex
\section{Conclusion}
We introduce the one-shot motion customization task that learns the motion signature from a single reference video and transfers it to new scenes and subjects with variety in both appearance and motion.
We propose Temporal LoRA to model the target motion by adding LoRA residual weights on the temporal attention layers of a pre-trained text-to-video diffusion model.
We further propose Appearance Absorbers to decouple the spatial information from the reference video so that Temporal LoRA can focus on motion modeling.
Extensive experiments demonstrate that our methods yield faithful and diverse videos compared to both per-frame video editing approaches and the base T2V model.
Our method's is plug-and-play nature supports various downstream tasks including precise video editing, video appearance customization, multiple motion combination as well as third-party appearance absorbers.

%% file: secs/X_suppl.tex
\section{Implementation Details}

\subsection{Bypassing Temporal Layers in T2V Models}

Many diffusion-based T2V models such as \cite{blattmann2023align, guo2024animatediff, wang2023modelscope, chai2023stablevideo} have their denoising network structure adapted from T2I UNet with temporal convolution and attention layers injected.
The new temporal layers are usually implemented as residual connections.
The models are also usually trained on image and video datasets jointly to acquire both appearance and motion generative capability.

Based on this mechanism, we propose to train our appearance absorbers with the temporal layers bypassed and the model to perform image generation tasks on static frames. 
This shared design further enables us to load third-party image customization models pre-trained on external image data to serve as ready appearance absorbers or additional spatial customization modules in our video applications.

\subsection{Patch Training of Appearance Absorbers}

Some motions are intrinsically highly associated with postures, such as walking, running and sitting, and one image can primarily represent them.
When the appearance absorbers have modeled the static postures to fit the appearance in the first training stage, T-LoRA might have little left to learn such as only the trivial perturbations across frames.

Therefore, we propose to crop the unordered frames into patches and encourage the appearance absorbers to mainly capture local shapes and textures in the first training stage.
This prevent our appearance absorbers from overfitting on the global structures fundamentally.
In practice, we find that setting the crop ratio randomly between $0.33$ to $0.67$ yields the best effect to retain the desired motion evidently in the second training stage.

\subsection{Attentions and LoRAs in T2V Diffusion Models}

\begin{figure}[t]
  \centering
  \includegraphics[width=\linewidth]{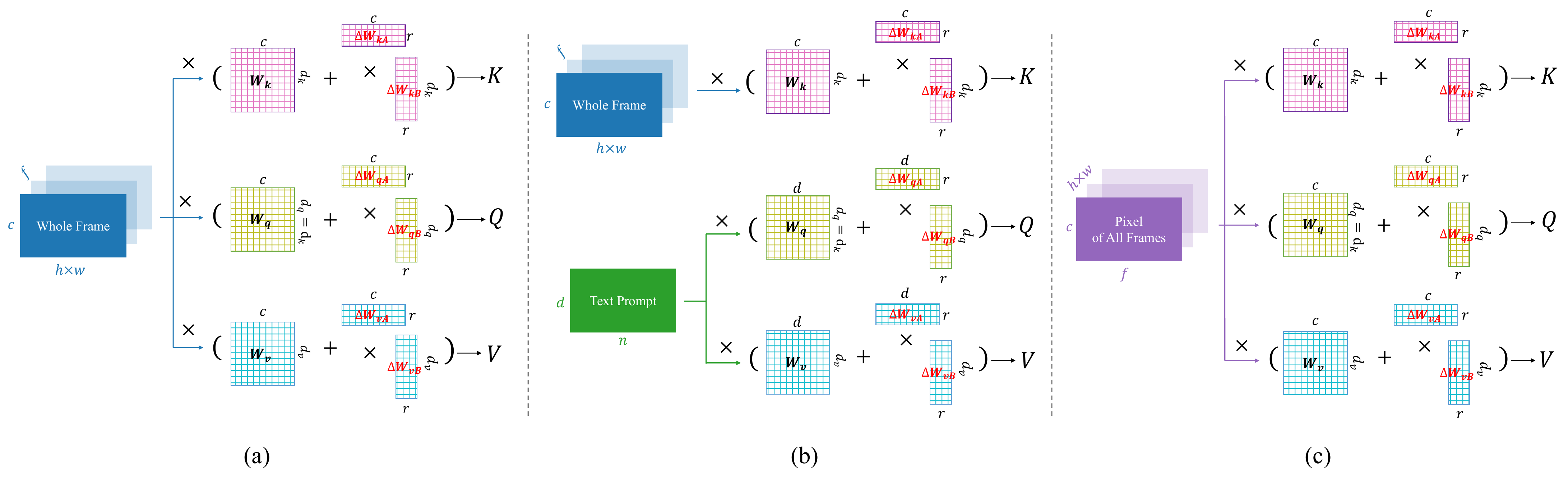}
   \caption{
   (a) Spatial self-attention between each frame and itself;
   (b) Spatial cross-attention between each frame and the text prompt;
   (c) Temporal cross-frame attention among pixels of all frames in a video.
   Batch size is omitted for simplicity.
   }
   \label{fig:attentions}
\end{figure}

A base T2V diffusion model involves spatial self-attention (SSA) between a frame and itself, spatial cross-attention (SCA) between each frame and the text prompt, and temporal cross-frame attention (TCFA) among a pixel across all time in each 3D UNet block.
We display their computations in Fig.~\ref{fig:attentions}.
Three types of input and their corresponding $K$, $Q$ and $V$ are marked in respective colors.
SSA is calculated between each frame and itself (\textcolor{blue}{$K$}, \textcolor{blue}{$Q$}, \textcolor{blue}{$V$}).
SCA is between a frame and the text prompt (\textcolor{blue}{$K$}, \textcolor{green}{$Q$}, \textcolor{green}{$V$}).
TCFA is among pixels of all frames (\textcolor{violet}{$K$}, \textcolor{violet}{$Q$}, \textcolor{violet}{$V$}).
Our LoRAs are applied to all attention weights $W_{*}$ (\textcolor{red}{$\Delta W_{k}$}, \textcolor{red}{$\Delta W_{q}$}, \textcolor{red}{$\Delta W_{v}$}).

\subsection{Model Hyperparameters and Training Time}

LoRA~\cite{hu2021lora} typically features very few additional parameters attached to the base model.
Its rank $r$ controls the shape of the residual matrix, and $\alpha$ represents its scale when added to the pre-trained model weights.
In experiments we discovered that setting the rank of T-LoRA $r_{T}=4$ and the rank of S-LoRA in the appearance absorber $r_{S}=1$ yields satisfactory results.
Meanwhile, we empirically determined the alpha values $\alpha_{T}=1$ for T-LoRAs and $\alpha_{S}=0.5$ for S-LoRAs.
For textual inversion~\cite{gal2022image} as the appearance absorber, we set the length of new learnable tokens to 2-6 depending on the content complexity.

\begin{table}
  \caption{
    Quantitative and model size comparison with concurrent work.
  }
  \centering
  \begin{tabular}{@{}lccccccc@{}}
    \toprule
    \multirow{2}{*}{Method} & Text & Temp. & \multirow{2}{*}{Div.$\uparrow$} \quad & \multicolumn{2}{c}{LoRA Rank} & \multicolumn{2}{c}{\#Params} \\
    \cmidrule{5-8}
    ~ & Align.$\uparrow$ \quad & Consist.$\downarrow$ \quad & ~ & \quad Temp. \quad & \quad Spat. \enskip & \enskip Temp. \enskip & \enskip Spat. \enskip \\
    \midrule
    Ours No AA & $31.687$ & $0.166$ & $0.613$ & \multirow{4}{*}{4} & - & \multirow{4}{*}{831.5K} & - \\
    Ours S-LoRA AA & $31.913$ & $0.163$ & $0.618$ & ~ & 1 & ~ & 207.5K \\
    Ours TextInv AA & $\boldsymbol{32.632}$ & $\boldsymbol{0.160}$ & $\underline{0.621}$ & ~ & - & ~ & 4K \\
    Ours Both AA & $32.193$ & $0.164$ & $\boldsymbol{0.631}$ & ~ & 1 & ~ & 211.5K \\
    \midrule
    MotionDirector~\cite{zhao2023motiondirector} & $31.842$ & $0.166$ & $0.595$ & 2 & 1 & 779.5K & 274.5K \\
    MotionDirector~\cite{zhao2023motiondirector} & $\underline{32.500}$ & $\underline{0.163}$ & $0.606$ & 4 & 1 & 1559K & 274.5K \\
    \bottomrule
  \end{tabular}
  \label{tab:concurrent}
\end{table}

We run experiments on a single NVIDIA RTX A5000 GPU with half-precision floats.
Our T-LoRA takes approximately 7 minutes to converge.
S-LoRA takes around 0.5 minute and the textual inversion takes 1 minute to converge in the first training stage.
This is comparable to Tune-A-Video~\cite{wu2023tune} (6 minutes), Video-P2P~\cite{liu2023video} (8 minutes in fast mode; 14 minutes in full mode on A6000 for bigger VRAM) and concurrent work MotionDirector~\cite{zhao2023motiondirector} (8 minutes) on the same device for the same frame resolution and clip length.
The learning rate is set to $5\times 10^{-4}$ for T-LoRA and $5\times 10^{-5}$ for appearance absorbers to prevent overfitting.

It worth noting that due to the difference in LoRA applications between our method and concurrent work MotionDirector, the quantities of parameters and module sizes are not aligned with the same LoRA rank.
We apply LoRAs on temporal cross-frame attentions (TCFAs), while MotionDirector moreoever add them to the following feed-forward networks (FFNs).
This lead to approximately twice the quantity of parameters to tune.
We apply LoRAs on all spatial attentions including the self-attentions (SSAs) and the cross-attentions (SCAs).
MotionDirector excludes the SCAs and additionally involves the following FFNs.
Thus our spatial LoRAs have comparable amounts of parameters.
Tab.~\ref{tab:concurrent} expands the quantitative comparison with these model size differences.

\section{More Visualizations}

\subsection{Video Generation Results}

\begin{figure}
  \centering
  \includegraphics[width=\linewidth]{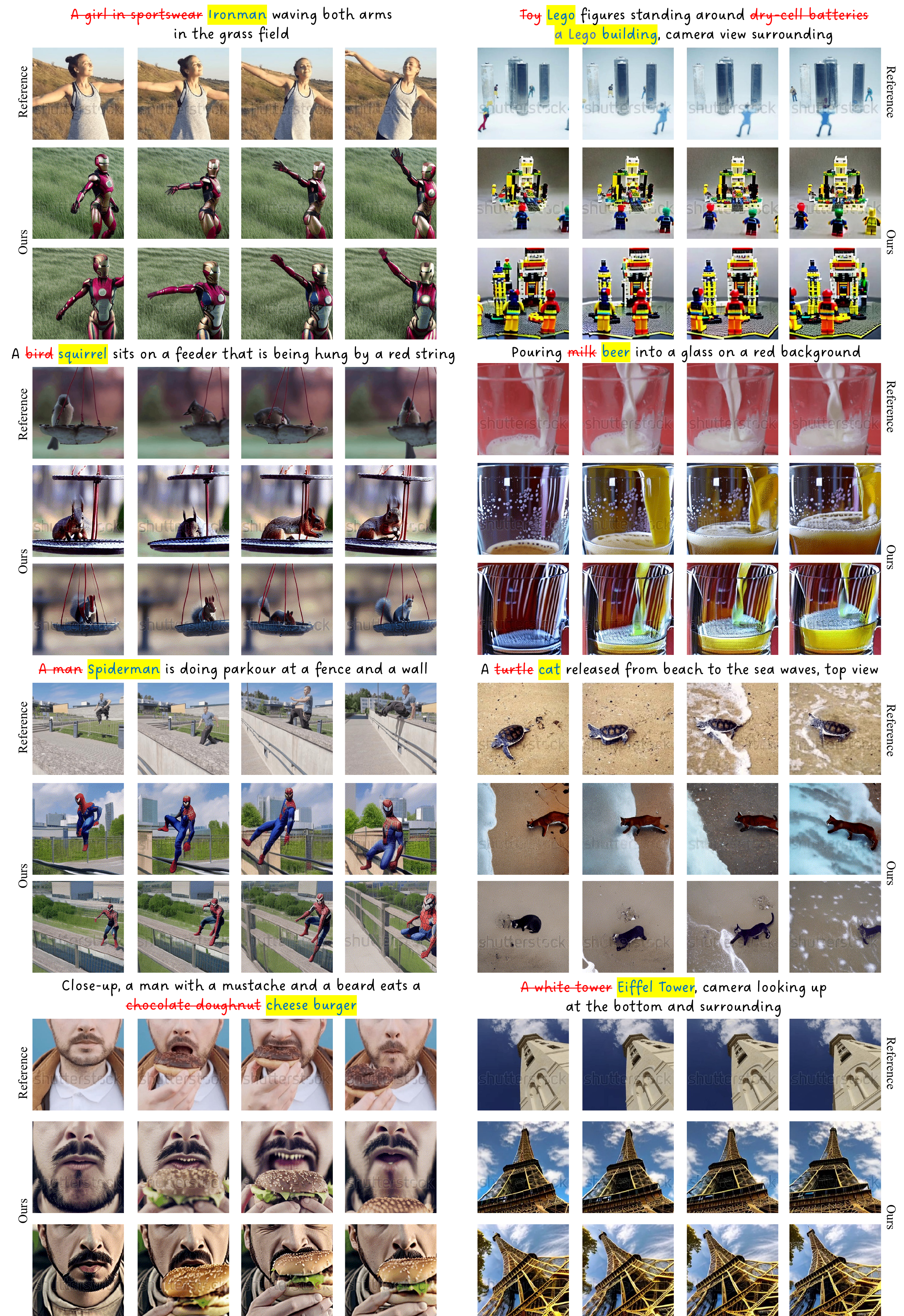}
  \caption{
    Additional generation results of our method.
  }
  \label{fig:extra}
\end{figure}

More video results generated by our models are displayed in Fig.~\ref{fig:extra}.
We present two random output samples for each reference video.

\subsection{Appearance Absorber Results}

\begin{figure}
  \centering
  \includegraphics[width=\linewidth]{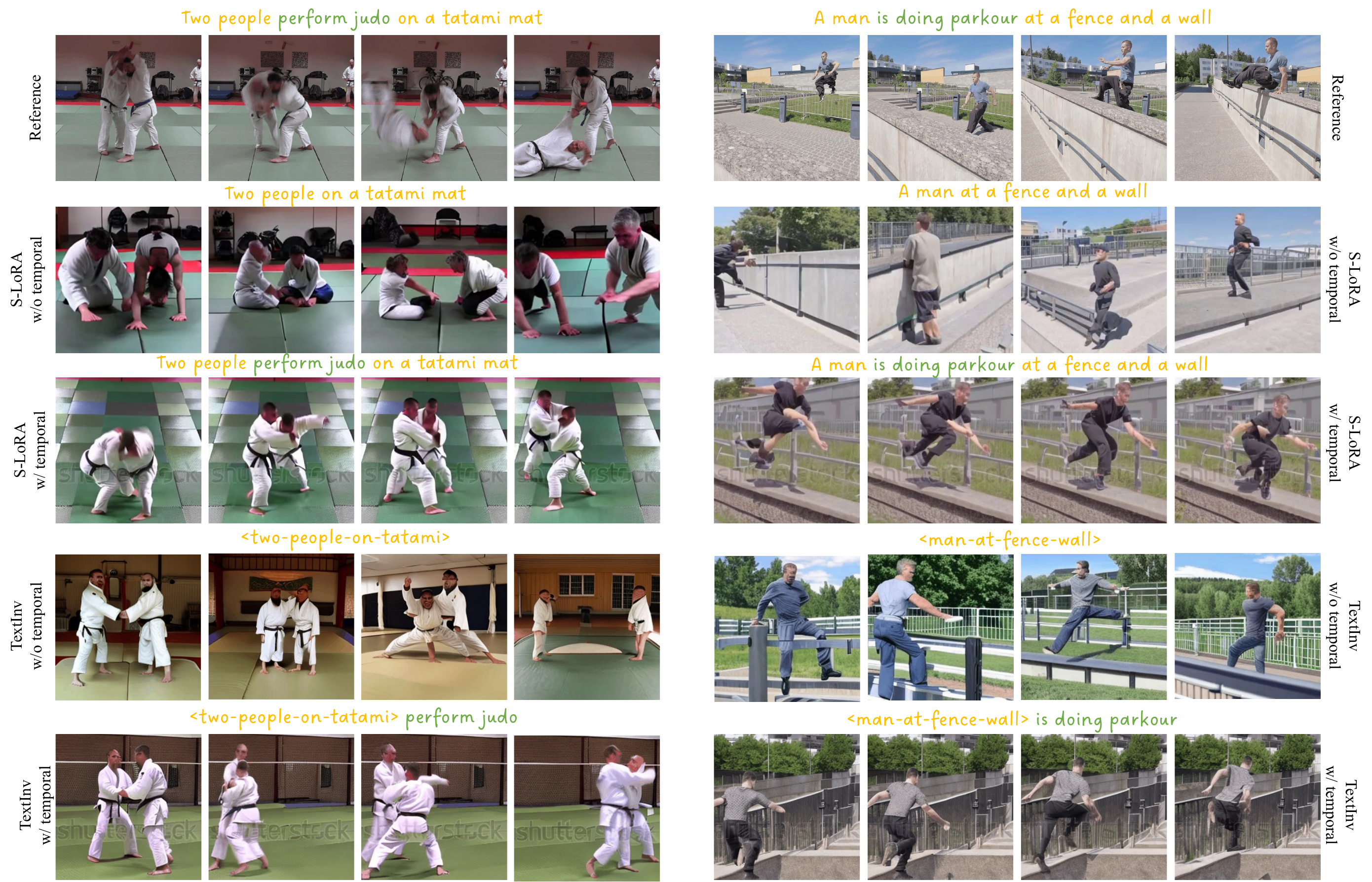}
  \caption{
    Appearance absorbers' generation output.
    The 2nd and 3rd rows have S-LoRA loaded.
    The 4th and 5th rows have textual inversion loaded.
    The training prompts and special tokens are noted above for each sample.
  }
  \label{fig:aa_output}
\end{figure}

We exhibit the output of our appearance absorbers trained on unorder reference frames with the spatial text prompt in Fig.~\ref{fig:aa_output}.
The 2nd and 4th rows show the generation results with the appearance absorbers (S-LoRA and textual inversion respectively, same below) loaded and all temporal layers bypassed in the base T2V model, and the spatial part of the text prompt is used.
It yields individual static frame replicating the reference appearances with random postures.
The dynamic information is successfully left for our temporal customization module to learn in the next stage.
The 3rd and 5th rows show the output videos with the appearance absorbers loaded on the full base T2V model, and the full text prompt is used.
With the temporal description, the model can still only produce generic motions upon the learned appearances, indicating the necessity and effectiveness of our temporal customization module training.
It can be further noticed that S-LoRA and textual inversion have different flavors of spatial modeling due to their different mechanisms, and thus loading both of them achieves the best performance with comprehensive and thorough appearance absorbing.

\subsection{Training Schedule Variances}
\label{sec:rvaa}

\begin{figure}
  \centering
  \includegraphics[width=\linewidth]{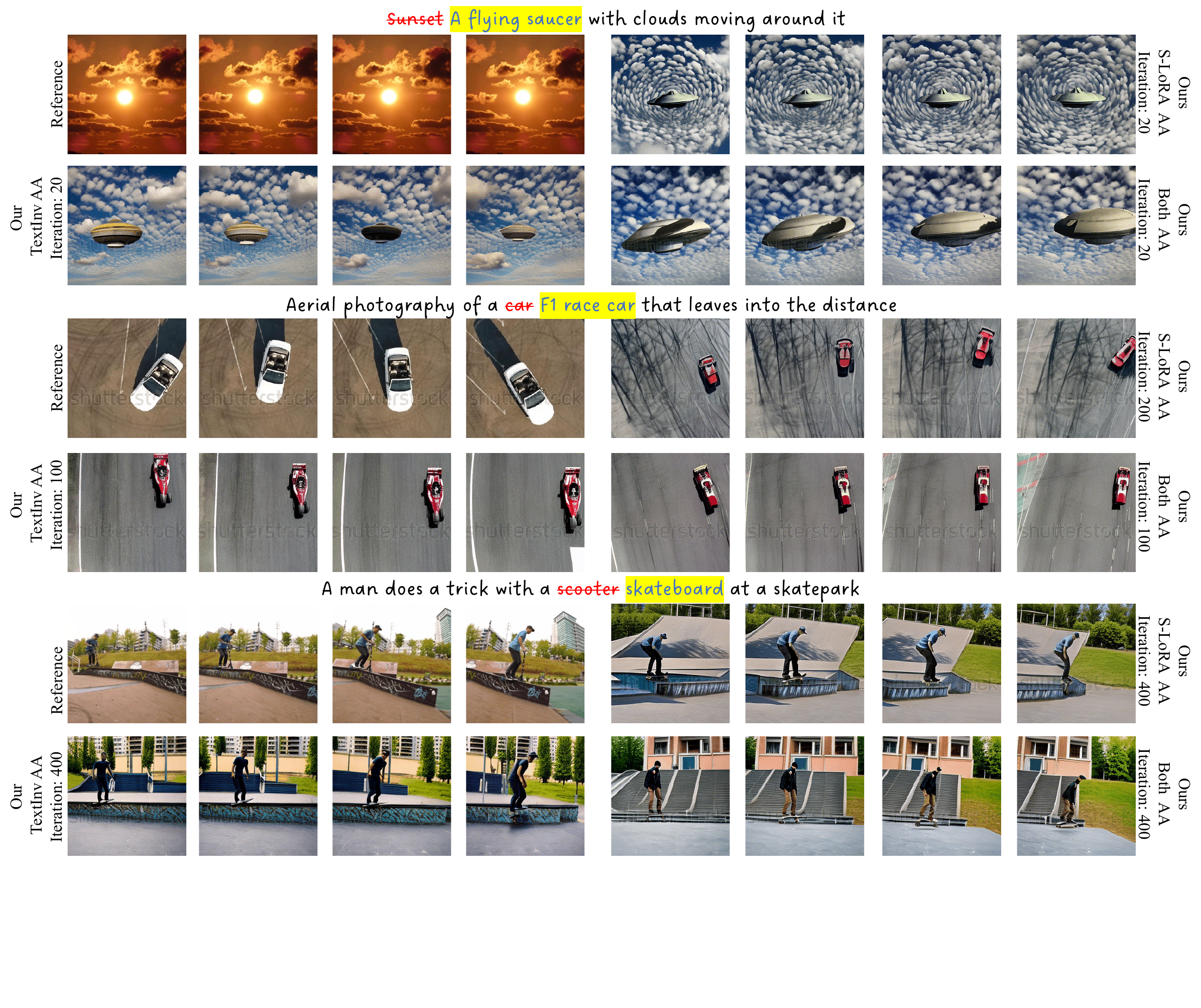}
  \caption{
    Examples where different reference videos require different tuning iterations.
    (1-2) Simpler motions such as camera movements usually converge faster.
    (3) More complicated motions such as animal or human actions would demand more tuning steps.
  }
  \label{fig:ablatAA1}
\end{figure}

Our modules fit on each reference video individually to model its unique motion signal.
Fig.~\ref{fig:ablatAA1} displays cases whose optimal iterations vary across different reference videos.
In general we observed that the convergence steps increase along with the complexity the specific reference motion and that of the original appearance.

\begin{figure}
  \centering
  \includegraphics[width=\linewidth]{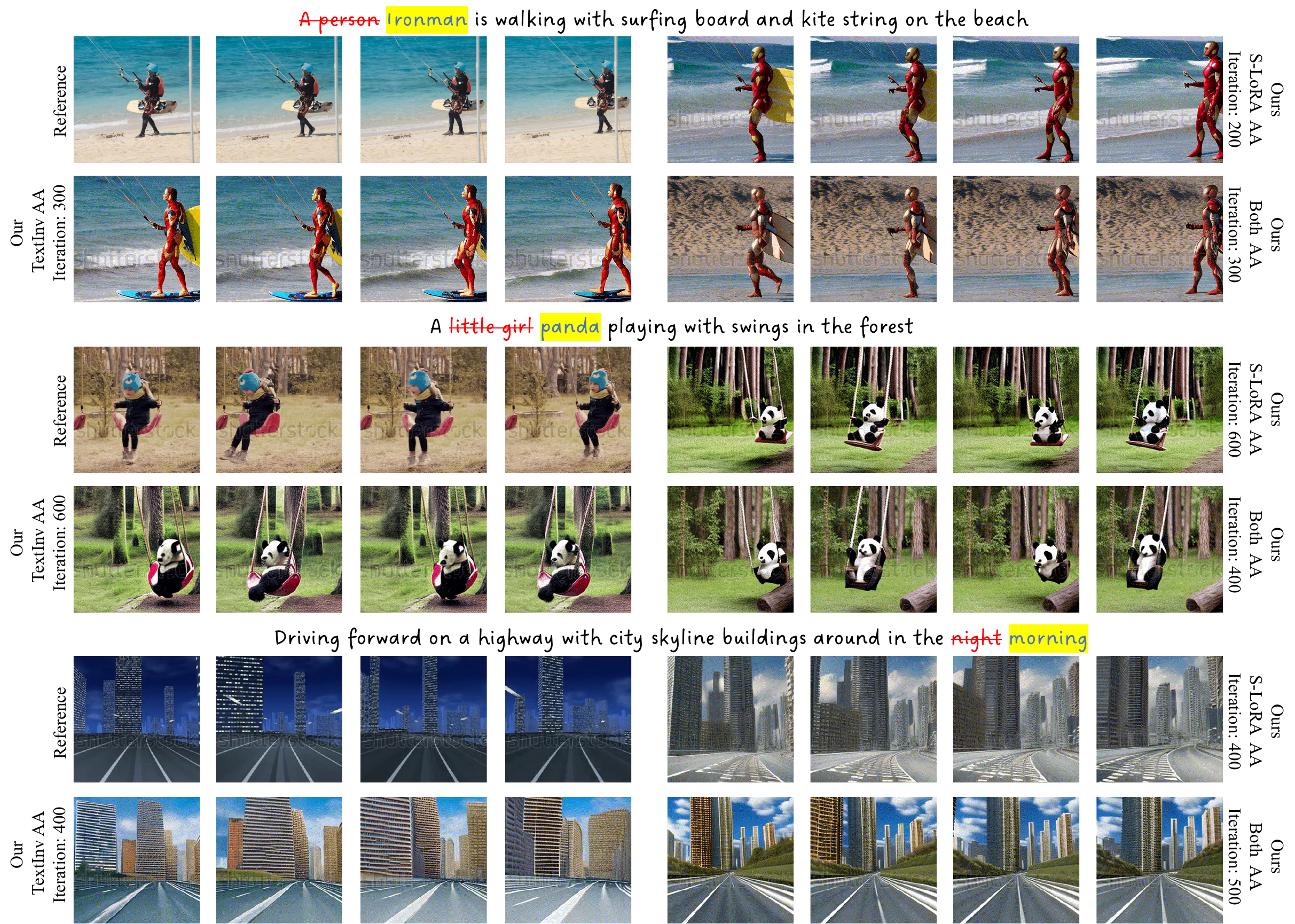}
  \caption{
    Examples where different appearance absorbers exhibit different characteristics.
    (1) Our Both AA absorbs the original appearance more thoroughly, leading to more diverse new background generated.
    (2) Our Both AA may reduce the necessary convergence step compared to a single appearance absorber.
    (3) Our Both AA may be more stable and enable more tuning iterations without collapse to thoroughly clean up the original art style and generate a new one.
  }
  \label{fig:ablatAA2}
\end{figure}

We also observed that different types of appearance absorbers may exhibit different characteristics that affect the optimal checkpoint step and the output details.
In Fig.~\ref{fig:ablatAA2} we present some cases where appearance absorbers vary in their effect of assisting following stages to capture the accurate motion or to generate novel scenes in certain iterations.

\section{User Study Questionnaire Design}

\begin{figure}
  \centering
  \includegraphics[width=\linewidth]{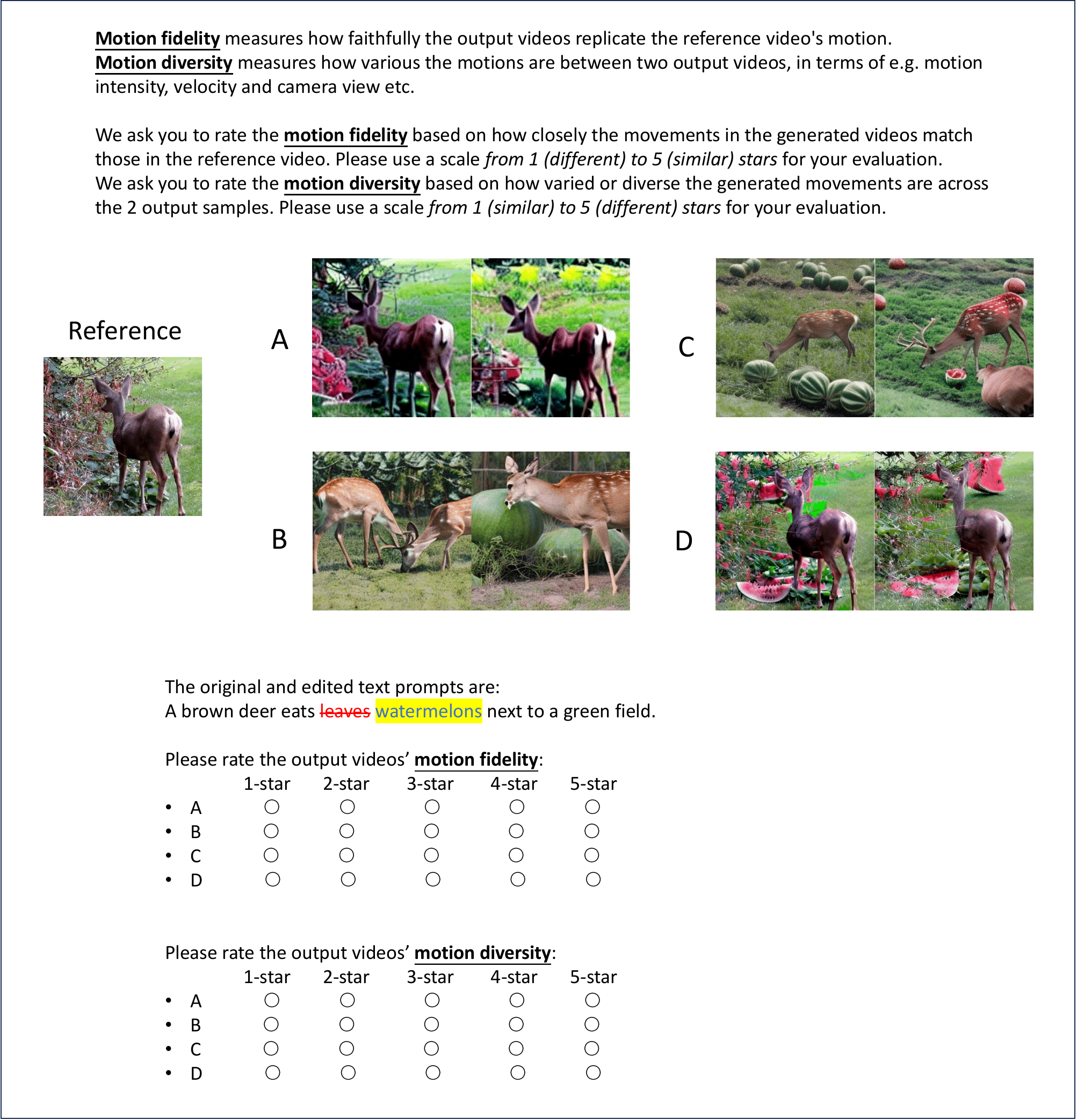}
  \caption{
    An example question in the human user study.
    Participants are asked to rate each algorithm's output videos from 1 to 5 stars.
  }
  \label{fig:us2}
\end{figure}

We present example questions in our user studies in Fig.~\ref{fig:us2}.
Every reference video is presented with the output videos by random 4 out of 5 algorithms to be evaluated.
For motion fidelity, 1-star represents the most dissimilar and 5-star represents the most faithful transferred motions w.r.t. the reference video.
For motion diversity, 1-star indicates the most identical and 5-star indicates the most diverse generated motions among the two output videos.

\section{Limitation Discussions}

\paragraph{Per Instance Finetuning.}
Our method tunes on each reference video individually.
Similar to comparing approaches~\cite{zhao2023motiondirector, wu2023tune, liu2023video}, our method needs specialized recipes for different videos of diverse appearances and motions.
The training configurations and iterations depend on the target video and can vary a lot as analysed in Sec.~\ref{sec:rvaa}.
The trade-off balance between the object motion fidelity and its diversity also relies on dedicated hyperparameters and adjustments between underfitting and overfitting, like its image customization counterparts~\cite{ruiz2023dreambooth, gal2022image} have described.
Though, our staged training pipeline and plug-and-play designs enable reusing both the appearance absorbers and T-LoRAs for future training and compositional inference, which improve their usability.

\paragraph{Spatial Domain Shift.}
The standalone finetuning of partial layers might have the risk of breaking the consistency among the pre-trained weights if the appearance absorbers overfit on static content reconstruction.
If the reference frames are out of the T2V model's pre-trained generalization capacity, the spatial customization might shift its output domains during training and the subsequent temporal layers will be unable to parse the altered feature maps properly in the next stage.
We suggest smaller learning rate and LoRA scale to pick the checkpoint when the reference video has complex appearances such as uncommon contents or extraordinary styles.
Applying our methods on advanced base T2V models with leading capabilities also helps.

\paragraph{Text Encoding Conflict.}
While extensive spatial customization modules can be alternatively utilized as our appearance absorbers, some of them might encounter text mapping conflict when collaborating with the temporal customization modules.
For example, we choose not to apply LoRA on the text encoder in T2V diffusion models although it can enhance the spatial modelling and appearance decomposition.
Modifications on the pre-trained text encoder could tamper the original mapping from text to its embedding, and then T-LoRA will learn the motion associated with the altered text tokens.
Finally it might not be triggered properly by the vanilla text encoder without the appearance absorbers during inference.
The null-text prompt training trick for LoRA without triggering words might help to handle this issue.

\section{Future Work}
Abundant image customization approaches with various tuning techniques have been developed for T2I diffusion models.
We leverage some of them to serve as our appearance absorbers for their training stability on few-shot learning and inference simplicity in the staged scheme.
In the next step we plan to investigate more options to discover their characteristics and further enhance our method's performance and usability.
Besides, generative video foundation models are also rapidly evolving and our modules are inherently compatible with various types of temporal attentions, regardless of the specific generation process and input modalities.